\documentclass[sigconf]{acmart}

\AtBeginDocument{%
  \providecommand\BibTeX{{%
    \normalfont B\kern-0.5em{\scshape i\kern-0.25em b}\kern-0.8em\TeX}}}

\setcopyright{acmcopyright}
\copyrightyear{2022}
\acmYear{2022}

\acmConference[EcomGen, SIGKDD]{}{2022}{Washington DC, USA}
\usepackage{multirow}
\usepackage[table]{colortbl}

\usepackage[colorinlistoftodos]{todonotes}

\definecolor{Gray}{gray}{0.8}
\begin{document}

\title{Answer Generation for Questions with Multiple Information Sources in E-Commerce}

\author{Anand A. Rajasekar}
\affiliation{%
  \institution{Flipkart}
  \city{Bengaluru}
  \country{India}
}
\email{anand.ar@flipkart.com}

\author{Nikesh Garera}
\affiliation{%
  \institution{Flipkart}
  \city{Bengaluru}
  \country{India}
}
\email{nikesh.garera@flipkart.com}


\begin{abstract}
E-commerce users routinely post questions on products they are interested in purchasing. Answering questions quickly enables users to make better purchase decisions. Given the sheer number, it is not possible to manually answer the questions. Automatic question answering systems can use product related information (specifications, reviews, similar questions posted) and answer questions at scale. However, using these sources poses two main challenges: (i) The presence of irrelevant information and (ii) the presence of ambiguity of sentiment present in reviews and similar questions. \\
We propose Multi Source Question Answering Pipeline (MSQAP) that utilizes the available information by separately performing relevancy and sentiment prediction before generating a response. We exploit user submitted answers and thus reduce the need for supervised text spans required for training the passage retrieval based models for Question Answering. Our relevancy prediction model (BERT-QA) outperforms all variants and has an improvement of \textbf{12.36\%} in F1 score compared to BERT-Base. MSQAP shows an average improvement of \textbf{37.43\%} in ROUGE and \textbf{198.31\%} in BLEU compared to the highest performing baseline (HSSC-q). Human evaluation shows an overall \textbf{14.55\%} accuracy improvement over our standalone generation model (T5-QA) indicating more accurate answers and our approach provides a better user experience for \textbf{79.7\%} of the questions compared to answer extraction.

\end{abstract}

\ccsdesc[500]{Computing methodologies~Natural language generation}

\keywords{Natural language generation, Question answering, Transformers}
\maketitle

\section{Introduction} \label{intro}
Automatic question answering systems that respond to product related questions have gained a lot of attention in recent years. Customers usually post questions to evaluate a product before purchasing it. 
Unless these questions are answered by a user who purchased it, they go unanswered. We observed that questions answered on the same day led to a conversion of \textbf{13.1\%} followed by a steady drop for every unanswered day and stood at \textbf{1.3\%} after day 3. 
In such cases, we can utilize the information present in various sources of the product such as reviews, specifications, and duplicate questions to aid us in automatically creating a response. This led to a variety of works in review and other sources driven answer generation \cite{oaag, rdag, pdrg}. \\
\begin{table*}[h]
    \caption{Example of answer generation dataset with candidates}
    \label{tab:d9}
    \begin{tabular}{|l |p{7 cm} | p{7 cm}|}
    \cline{2-3}
\multicolumn{1}{c|}{}& Sample 1 & Sample 2 \\
\hline
Question & display are very slow in \textit{ABC}? & does phn have theatre sound quality?\\
Reference Answer & my mobile display slow & Yes, the audio quality of this phone is too good\\
\hline
Duplicate Q\&A (Partial) & 1) At present now on words which is better \textit{XYZ} or \textit{ABC} And mainly display which is better? \textit{ABC}. & 1) how is the sound quality while playing vides? Sound quality is superb. \\
\hline
Reviews & 1) \textit{ABC} mobile overall good but display quality poor & 1) Good enough to enjoy music without headphones. \\ 
& 2) but the display of \textit{ABC} are not good & 2) Sound quality is not good \\
\hline
Specifications &  1) Other Display Features: ...Narrow Frame: 2.05mm, Screen Ratio: &  1) Sound Enhancements: ...Noise Reduction: Dual-microphone Noise \\ 
\hline
    \end{tabular}
\end{table*}
\indent One of the primary challenges present in building answer generation systems for E-commerce is the noise present in the dataset, which consists of user posted questions and answers. These have spelling errors, grammatical inconsistencies, and sometimes code switching. Another common problem is the presence of irrelevant information and ambiguity of sentiment in users' opinions present in reviews and similar questions. Sample question, answer pairs along with their information candidates are tabulated in Table \ref{tab:d9}. Sample 2 contains reviews with contradicting sentiments for the same question. We attempt to address these challenges in our work. \\
\indent Transformers \citep{vas} have gained extensive popularity due to their top performance in a variety of NLP tasks. 
Transformers scale well with larger training data and allows efficient parallel training. In recent years, it is common to pre-train the transformer on a data rich task. This allow the model to learn general knowledge about the language that can be transferred to downstream tasks with a few steps of fine tuning. The initial pre-training step is often done in an unsupervised fashion on unlabelled data and has resulted in state of the art results in many NLP benchmarks \citep{bert, xlnet, albert}. The main advantage of this way of pretraining is due to the availability of large volumes of text data. We use pre-trained transformers as our models that are further finetuned for the task at hand.\\
\indent We propose Multi Source Question Answering Pipeline (MSQAP) consisting of three components, (i) relevancy prediction using a transformer fine tuned on Next Sentence Prediction task, (ii) sentiment prediction using a pre-trained model and (iii) answer generation using a text to text transformer fine tuned on a large Question Answer dataset to generate accurate and precise responses.

Our main contributions are:
\begin{itemize}
    \item An answer generation system for generating a natural language answer utilizing relevant extractions from reviews, similar questions and specifications. 
    \item An approach to handle the presence of (i) irrelevant information and (ii) ambiguity of answer sentiment,
    \item A novel pipeline (MSQAP) 
    that performs relevancy and sentiment prediction before generating a response. 
\end{itemize}


\section{Related work} \label{rel}
\noindent BERT \cite{bert} is a bidirectional transformer encoder pre-trained on large amounts of data to perform masked language modeling and Next Sentence Prediction task. RoBERTa \cite{roberta} removes NSP task from BERT's pretraining and introduces dynamic masking. 
T5 \citep{t5} is an encoder-decoder transformer architecture trained on a variety of tasks using transfer learning strategies. We employ all the above transformers in our study. 

\subsection{Relevancy prediction}
Recent studies \cite{aqa, dedupe} employ ranking strategies to pick an answer from a set of candidate answers while incorporating review information. \citet[]{dedupe} proposed a model to retrieve the most similar question from a list of QA pairs and used the corresponding answer as the response. \citet[]{cui} built a chatbot called SuperAgent, which extracts the information from all three sources through different approaches and prioritizes the results using a meta engine. \citet[]{productQnA} utilize all three sources of information to rank the candidates and pick the best answer that passes the relevancy threshold. 
\citet[]{QAR-net} posed this task as a binary classification problem predicting whether or not a review sentence answers the question. 
\citet[]{amaz} use transformer based models to retrieve question and answer pairs based on their relevance to the question. We use a similar approach to  select the top $k$ candidates and further add a key component called sentiment filtering given the inconsistency in answers from e-commerce user data. These are then used as context for our language generation model.
\subsection{Answer generation}
RNN \cite{oryl} and Transformer based \cite{bart, t5} models have shown promising results in text generation. These methods use attention mechanism \cite{nmt} and/or self attention \citep{vas}. 
\citet{oaag, rdag} use a RNN based model, with source information obtained from reviews. \citet[]{mcau} use a Mixture Of Expert model with review as the source to predict the answer, where they classify them into binary(Yes/No) answers. \citet[]{rdag} utilize  attention mechanism \cite{rdag} to alleviate the noise present in review snippets. \citet{dish} use BERT \cite{bert} on reviews to answer binary questions. \citet[]{oaag} learn a multi task model that performs answer generation and opinion mining while utilizing review ratings. They use a pointer generator network \cite{pg} with fusion of review information to generate answers. \citet[]{pdrg} encode reviews and product specifications using separate encoders. The generated answer is passed through a consistency discriminator to check if it matches the facts. \citet[]{mag} generate a coherent response using the information present in reviews and product descriptions and a prototype answer as a template. Our focus in this work is to build a comprehensive answer generation pipeline utilizing information from various e-commerce sources such as specifications, similar questions, and reviews. To the best of our knowledge, this is the first work in e-commerce domain to utilize all the above three sources of information for answer generation. We utilize user submitted answers for answer generation and thus reducing the need for clean supervised data present in most of the prior work.

\section{Problem Definition} \label{prob}

Given a question $Q$ and a set of information candidates $\{x_1, \dots, x_k\}$ related to a product, the goal is to generate a natural language answer $y$ as the response using relevant information.

Our dataset $D$ consists of $N$ samples where data point, $d^{i}$ consists of the question $Q^i$, a set of reviews $\{r^i_1, \dots, r^i_k\}$, a set of duplicate questions and answers $\{(q^i_1, a^i_1), \dots, (q^i_l, a^i_l)\}$, and a set of specifications $\{s^i_1, \dots, s^i_m\}$ and the ground truth answer $y^i$. Formally,  

\begin{equation}
D = \big( Q^i, \{r^i_1, \dots, r^i_k\}, \{(q^i_1, a^i_1), \dots, \\ (q^i_l, a^i_l)\}, \{s^i_1, \dots, s^i_m\}, y^i\big)_{i=1}^N    
\end{equation} 

\noindent The goal is to generate a coherent and precise answer $\hat{y}^i$ using appropriate information.


\section{Proposed approach} \label{prop}
\begin{figure}[h]
  \centering
  \includegraphics[width=\linewidth]{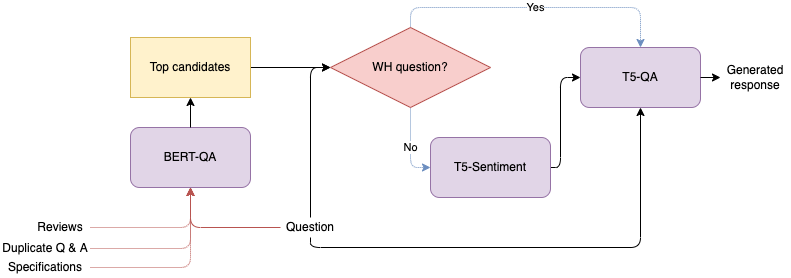}
  \caption{Multi Source Question Answering Pipeline}
  \label{fig:msqap}
\end{figure}

\noindent We introduce MSQAP to generate the answer to a product related question by removing irrelevant information and ambiguity of sentiment present in the information candidates. Figure \ref{fig:msqap} depicts the overview of MSQAP consisting of three components. $(1)$ Relevancy prediction using relational information between the question and candidates and ranking their importance in answering a question. $(2)$ Sentiment prediction to remove ambiguous opinions present in the candidates. $(3)$ Answer generator to generate the response to the question using relevant context. The models used in this pipeline are trained and evaluated separately.

\subsection{Relevancy prediction}
The candidates available in the dataset are quite high in number for common questions and fewer in number for unique/rare questions. Also, these candidates are sometimes repetitive or irrelevant in their ability to answer the query. Using many candidates for generation will increase computation time and will make it difficult for the model to give attention to the right information. Hence, it is important to rank them with respect to their relevance to the question and their importance in answering it. We propose to model relevancy prediction as a Next Sentence Prediction task with the question as the first sentence and each candidate as the potential second sentence. 
The intuition behind this modeling strategy is that any candidate that is relevant for answering a question will be the sentence immediately following it in a paragraph. Any other sentence that is irrelevant will be a random sentence. Thus, a well-trained NSP model should be able to identify an important and relevant candidate from a random one.

Let $n^i$ be the number of candidates available for question $Q^i$. The question $Q^i$ and a candidate $x^i_j$ are concatenated, tokenized, and passed to an embedding layer. The word embeddings along with their positional signals are passed to a transformer encoder whose head predicts the Next Sentence label. 

\begin{equation}
w^i_j = tokenizer([Q^i;x^i_j]) \;\;\;\;
\hat{y}^i_j = transformer(w^i_j)
\end{equation} 

\noindent $";"$ denote the appropriate concatenation of input sentences as required by the pre-trained transformer in use. The relevancy prediction task is trained to minimize the binary cross entropy loss,
\begin{equation}
L_{nsp} = -\frac{1}{N_1} \sum_{i=1}^{N_1} \frac{1}{n^i} \sum_{j=1}^{n^i} y^i_j log \hat{y}^i_j
\end{equation} 
where $y^i_j$ is the ground truth label indicating the relevancy of the candidate and $N_1$ refers to the number of data points in Dataset $D1$. 
\subsection{Sentiment prediction}
The other reason for unnecessary information being fed into the generation model is the ambiguity present in the candidates due to the polarity of sentiment.  When both positive and negative sentiments together are passed as inputs to the generation model during training with a ground truth label containing any of the two sentiments, it will hinder the model from learning to generate a response with the right sentiment. For eg., two data points with the same question and candidates but using conflicting answers given by two users as ground truth labels will confuse the model whereas filtering the candidates to match the sentiment of the label will help the model to generate a response with sentiment present in the input.  Hence, we have to remove candidates with opposite sentiments to that of the label and the advantage of removing this ambiguity is twofold. First, the number of input candidates decreases improving the computation time, and second, the model gets trained to generate an answer with the sentiment given as input. Sample 2 in Table \ref{tab:d9} contains both positive and negative sentiments. We remove the second review since it contradicts the reference answer before creating the context for training. If the reference answer was negative, we would remove candidates with positive sentiment. During evaluation, we choose the sentiment that is in majority among the candidates and the generated answer will contain the chosen sentiment. However, this task is carried out only for dichotomous (yes/no) questions. The candidates of WH questions (What, When, Where etc.,) usually have less polarity in sentiment and hence, they are left untouched.  

\subsection{Answer generation}
Once the candidates that are relevant to answering a question are collected, they are concatenated and passed along with the question into a transformer with encoder decoder architecture to generate the response. The encoder-decoder implementation follows the original proposed form \citep{vas}. The encoder consists of a stack of layers each with a self attention layer and a feed forward network. Layer normalization is applied to input of every layer and a skip connection is applied which connects the layer's input with its output. The decoder consists of a similar setup except that it also has an attention layer that attends to the output of the encoder. The self attention layer in decoder follows a causal attention strategy (i.e) paying attention to past inputs only. \\
\indent The number of filtered candidates available for question $Q^i$ is $\leq k$. The question $Q^i$ and its candidates are concatenated, tokenized, and passed to an embedding layer. The word embeddings and their positional signals are passed to a transformer encoder which helps the decoder in generating the answers in an auto regressive fashion.

\begin{equation}
\begin{aligned}
w^i = tokenizer([Q^i;(x^i_1:\dots:x^i_{\leq k})])  \\
\hat{h}^i = Encoder(w^i) \;\;\;\; \hat{y}^i_t = Decoder(h^i, y^i_{t-1}) 
\end{aligned}
\end{equation} 

\noindent $":"$ denote the regular string concatenation and $\hat{y}^i_t$ is the generated token at position $t$. The answer generation task is trained to minimize the cross entropy loss,

\begin{equation}
L_{gen} = - \frac{1}{N_2} \sum_{i=1}^{N_2} log P(\hat{y}^i_*)
\end{equation} 

\noindent where $P(\hat{y}^i_*)$ is the token probability corresponding to ground truth and $N_2$ indicates the number of data points in Dataset $D2$.
\subsection{Auto answering pipeline}
The MSQA Pipeline brings together all three components in aiding answer generation. First, all the provided candidates from reviews, similar questions and answers, and specifications are passed as inputs to the relevancy prediction task. The scores obtained from the NSP model are used to rank the candidates in decreasing order of their relevancy. The top $k$ candidates are chosen and passed to the sentiment prediction task depending on the type of question. The trimmed candidates along with the question are passed as input to the answer generation model to produce the response. 

\section{Experiments}\label{exp}
The aim of the analysis is to answer the following questions
\begin{itemize}
\item Does the pipeline outperform the results of the baselines?
\item Are the generated answers precise and coherent?
\item How does each variant of the pipeline generate answers?
\item How does generation compare to retrieved answers?

\end{itemize}
\subsection{Dataset}
We train and evaluate the relevancy prediction model using our inhouse dataset $D1$ of mobiles. This dataset is made up of questions collected randomly from a list of accepted user posted questions. The number of questions present in the dataset is 2000. Every question is matched with a set of candidates from three sources of information, reviews, similar questions and answers, and specifications. We manually labeled this dataset to indicate whether each candidate has relevant information to answer the question.
\begin{table}[h]
  \caption{Relevancy prediction dataset}
  \label{tab:d1}
  \begin{tabular}{|l|c|c|}
    \cline{2-3}
    \multicolumn{1}{c|}{} &Train dataset &Test dataset\\
    \hline
    Total candidates &  15122 & 3268\\
    Total relevant candidates &  8670 & 1736\\
    Avg. specs relevancy & 0.308 & 0.253\\
    Avg. qa relevancy & 0.668 & 0.634\\
    Avg. reviews relevancy & 0.626 & 0.573\\
  \hline
\end{tabular}
\end{table}

\noindent Specifically, dataset $D1$ is represented by,
\begin{align*} 
D1 = \big(Q^i,\{(r^i_1, yr^i_1), \dots, (r^i_k, yr^i_k)\}, \{(q^i_1, a^i_1, ya^i_1), \dots, \\ (q^i_l, a^i_l, ya^i_l)\}, \{(s^i_1, ys^i_1), \dots, (s^i_m, ys^i_m)\}\big)_{i=1}^N
\end{align*}
\noindent where $yr^i_j$, $ys^i_j$, and $ya^i_j$ denote the relevancy of each of the sources for $j^{th}$ candidate of $i^{th}$ question. Every question with any one of the candidate makes up a datapoint for the Next Sentence Prediction task. The statistics of the dataset are presented in Table \ref{tab:d1}. The dataset is well balanced with an average of 0.573 and 0.531 relevant candidates in the train dataset and test dataset respectively. \\
\vspace{-1.2em}
\begin{table}[h]
  \caption{Answer generation dataset}
  \label{tab:d2}
  \begin{tabular}{|l|c|c|}
    \cline{2-3}
    \multicolumn{1}{c|}{}&Train dataset &Test dataset\\
    \hline
    Total No. of questions &  217086 & 5423\\
    No. of WH questions &  66075 & 2031\\
    Avg. candidates per question & 9.750& 10.55\\
    Avg. specs per question &  2.381 & 2.429\\
    Avg. reviews per question &  2.637& 2.553\\
    Avg. dup. questions per question &  4.732 & 5.568\\
    
  \hline
\end{tabular}
\end{table}

\begin{table}[h]
\caption{Datasets used for training models}
\begin{center}
\begin{tabular}{|c|c|c|}
    \hline
    Dataset & $D1$ & $D2$ \\
    \hline
    \multirow{4}{*}{Models}&RoBERTa-A& Seq2Seq\\
    &BERT-A & HSSC-q\\
    &RoBERTa-QA & T5-QA\\
    &BERT-QA&\\
    \hline
\end{tabular}
\end{center}
\label{tab:data}
\end{table}

\noindent We train and evaluate the answer generation model using our inhouse dataset $D2$ of mobiles consisting of 200K questions collected randomly from user posted questions along with their answers. The ground truth labels of this dataset are noisy since they represent the individual opinion of a single user. We also collect candidates from all three sources for each question. Specifically, dataset $D2$ is represented by,
\begin{align*} 
 D2 = \big( Q^i, \{x^i_1, \dots, x^i_{n^i}\}, y^i\big)_{i=1}^N
\end{align*}

\noindent The statistics of the dataset are presented in Table \ref{tab:d2}. Though the duplicate questions and answers are more relevant in answering the question as evident from Table \ref{tab:d1}, they are also more in number per question on average. The datasets used for training our models are tabulated in Table \ref{tab:data}. 

\subsection{Baselines \& Evaluation metrics}
T5-QA model denotes answer generation component only, while MSQAP (rel.) denotes our generation model with relevancy prediction and MSQAP denotes the entire pipeline. We compare our approach on both relevancy prediction and answer generation baselines. We have adopted two generation based methods along with the pre-trained base transformer in use for answer generation task.

\begin{itemize}
    \item \textbf{Seq2Seq \citep{nmt}} - We implement the standard sequence to sequence RNN model with attention. The question and the candidates are concatenated and fed as input to the model. \item \textbf{HSSC-q} - We utilize the multi task model HSSC \cite{hssc} that jointly performs summarization and sentiment classification. However, we implement a slightly modified variant HSSC-q that utilizes question with candidates to perform answer generation and sentiment classification of generated answer. 
    \item \textbf{T5-Base \citep{t5}} - We use the pre-trained text-to-text transformer trained on a variety of tasks to do answer generation.

\end{itemize}
\begin{table}
  \caption{Methods comparison on relevancy prediction}
  \label{tab:d4}
  \begin{tabular}{|l|c|c|c|c|}
    \hline
    Model & Acc & Pre & Rec & F1  \\
    \hline
    BERT-base & 0.635& 0.637& \textbf{0.996} & 0.777\\
    \hline
    RoBERTa-A & 0.708 &0.767 &	0.7778 & 0.772\\
    BERT-A & 0.749 &0.806 & 0.797 & 0.802\\
    RoBERTa-QA & 0.764 & 0.832 & 0.789 & 0.810\\
    BERT-QA& \textbf{0.838} & \textbf{0.873}& 0.872 & \textbf{0.873}\\
    \hline
  \end{tabular}
\end{table}

\begin{table*}[h]
  \caption{Evaluation of methods on answer generation}
  \label{tab:d5}
  \begin{tabular}{|*{2}{c|}*{4}{c}|*{4}{c}|*{4}{c}|}
    \cline{3-14}
    \multicolumn{1}{c}{}&\multicolumn{1}{c}{} & \multicolumn{4}{|c|}{Dichot. questions} & \multicolumn{4}{c|}{WH questions} & \multicolumn{4}{c|}{Overall dataset} \\
    \cline{3-14}
    \multicolumn{1}{c}{}&\multicolumn{1}{c|}{}& R1& R2& RL& B1& R1& R2& RL& B1 &R1& R2& RL& B1 \\
    \hline
    \multirow{3}{*}{Baselines} & T5-Base & 11.88& 2.45& 11.18& 0.29& 10.01 & 2.86& 9.63& 0.14& 11.21& 2.61& 10.6& 0.22\\
    &Seq2Seq & 25.7 & 7.85& 24.79& 1.8&16.19 & 4.41 & 15.29 & 0.89& 22.14& 6.55& 21.25& 1.41\\
    &HSSC-q & 27.0& 10.19& 26.22& 2.4& 17.45 & 5.96 & 16.62 & 1.01& \cellcolor{Gray!25} 23.45& \cellcolor{Gray!25}8.61& \cellcolor{Gray!25}22.62& \cellcolor{Gray!25}1.78\\
    \hline
    \multirow{3}{*}{Ours} & T5-QA & 34.56& 14.98 & 32.8& 6.36& 24.63& \textbf{9.89} & 22.45& 3.65& 30.86& \textbf{13.09}& 28.93& 5.22\\
    &MSQAP (rel.) &  34.55& \textbf{15.1}& \textbf{32.89}& 6.4 & 24.32 & 9.49& 22.35& 3.56& 30.72& 13.01& 28.94& 5.21\\
    &MSQAP & \textbf{34.63}& 15.02& 32.88& \textbf{6.46} & \textbf{24.67} & 9.85& \textbf{22.68}& \textbf{3.72}& \cellcolor{Gray!25}\textbf{30.9}& \cellcolor{Gray!25}13.08& \cellcolor{Gray!25}\textbf{29.09}& \cellcolor{Gray!25}\textbf{5.31}\\
    \hline
    \multicolumn{1}{c}{}&\multicolumn{1}{c}{} & \multicolumn{4}{c}{} && \multicolumn{3}{c|}{\% improvement} & 31.77 & 51.92 & 28.6 & 198.31\\
    \cline{11-14}
  \end{tabular}
\end{table*}

\noindent We use a pre-trained transformer as the baseline for our relevancy prediction task. 

\begin{itemize}
    \item \textbf{BERT-Base \citep{bert}} - We compare our models with different architectural and input variants with the pre-trained BERT model on Next Sentence Prediction task. 

\end{itemize}

\noindent We use ROUGE (R1, R2, RL) and BLEU (B1) to automatically evaluate the performance of our answer generation pipeline. We also employ human evaluation of generated answers and measure Correct w.r.t candidates (CC) and Correct w.r.t label (CL) in order to quantify the  performance of our pipeline. We use Accuracy (Acc), Precision (Pre), Recall (Rec) and F1-score (F1) to evaluate the variants of our relevancy prediction model.

\begin{table}[h]
    \caption{Case study of generated answers}
    \label{tab:d7}
    \begin{tabular}{|p{8 cm} |}
    \hline

\textbf{Question 1: } what type of headset will i get in the box?\\
\textbf{Reference Answer: } No headset in the box. \\
\hline
\textbf{T5-base: } headset \\
\textbf{Seq2Seq: } no , it does n't contain any headset \\
\textbf{HSSC-q: } no headset in the box \\
\hline
\textbf{T5-QA: } No headset in the box \\
\textbf{MSQAP (rel.): } No headset in the box \\
\textbf{MSQAP: } No, there is no headset in the box.\\
 \hline
 \hline
\textbf{Question 2:} sound quality and heating problem?\\
\textbf{Reference Answer:} no issue at all and sound is great \\
\hline
\textbf{T5-base:} Hang \\
\textbf{Seq2Seq:} sound quality is good \\
\textbf{HSSC-q:} sound quality is good \\
\hline
\textbf{T5-QA:} sound quality is good but heating problem is there\\
\textbf{MSQAP (rel.):} sound quality is good but heating problem is there \\
\textbf{MSQAP:} sound quality is good and no heating problem \\
\hline
\hline
\textbf{Question 3:} is there 3gb ram version avilable?\\
\textbf{Reference Answer:} no never .4GB ram please \\
		
\hline
\textbf{T5-base:} 4 GB \\
\textbf{Seq2Seq:} no , it has a gb ram \\
\textbf{HSSC-q:} yes it is available \\
\hline
\textbf{T5-QA:} No, 4gb ram variant is not available.\\
\textbf{MSQAP (rel.):} No, 4gb ram variant is not available. \\
\textbf{MSQAP:} No, it's 4gb ram\\
\hline
    \end{tabular}
\end{table}

\subsection{Implementation details}
We use Transformers \cite{hug} package for loading and training pre-trained models. All models are implemented in Pytorch \cite{torch}. 

\subsubsection{\textbf{Relevancy prediction}}
We finetune BERT \citep{bert} model on dataset $D1$. We also finetune RoBERTa \citep{roberta} model on dataset $D1$ pre-trained on inhouse reviews data of mobiles vertical to perform relevancy prediction. We try two variants of these models by changing the input. The first variant utilizes only the duplicate question's answer as second sentence for prediction and is denoted by BERT-A and RoBERTa-A while the second variant uses both duplicate question and answer for prediction and is denoted by BERT-QA and RoBERTa-QA. The remaining two sources remain the same across variants. Every sentence in a review is considered as an individual candidate and key-value pair present in specifications is utilized directly(not converted into a sentence). All four models are trained for 5 epochs with a batch size of 32. 

\subsubsection{\textbf{Sentiment prediction}}
We use pre-trained T5 (T5-Sentiment) \cite{t5} model for sentiment prediction. T5-Sentiment had a good F1-score of \textbf{0.978} on an inhouse labeled data of 500 sentences. During training, we filter out the sentiments that are in contrast with the label and during evaluation, we keep the sentiment that is expressed by most of the candidates (i.e) minority sentiments are eliminated. Around one-third of the dataset contains WH questions as reported in Table \ref{tab:d2} and sentiment filtering is not performed on those points.

\subsubsection{\textbf{Answer generation}}
We train Seq2Seq model with pre-trained Glove embeddings \cite{glove} with 300 dimensions and with a vocabulary size of 400k. We train HSSC-q \cite{hssc} using the details provided in the paper. We finetune pre-trained T5 \citep{t5} model. We concatenate all the candidates from reviews, and specifications, and duplicate question \& answer pairs to a single sentence which is used as context for the models. Every duplicate question is followed by its corresponding answer and unnecessary punctuation is removed from the specifications. These candidates are filtered by relevancy/sentiment prediction models before training, depending on the pipeline variant. The number of top candidates ($k$) for MSQAP(rel.) and MSQAP is set as 7. Higher values of $k$ ($k$>7) does not improve performance but increases computation time. We train all the models with a batch size of 32. We set a learning rate of $5 \times 10^{-5}$.  We train all the models for 25 epochs. 

\subsection{Results}

\subsubsection{\textbf{Relevancy prediction}}
The relevancy prediction results are reported in Table \ref{tab:d4}. Though relevancy prediction is not directly related to answer generation, ranking the candidates based on their importance and picking the top $k$ candidates aids in the performance of generation. Our baseline model, BERT predicts almost all candidates as relevant and hence has a high recall. However, the precision of the baseline model is quite low proving that its ability to pick the relevant candidates is lower. Both the variants finetuned from BERT perform better than their counterparts owing to BERT being pre-trained on NSP task. Our models, BERT-A and RoBERTa-A have a moderate performance compared to QA variants due to the lack of duplicate question information. BERT-QA has the best performance because it combines the best of both worlds; NSP pretraining and QA information. We choose BERT-QA as the relevancy prediction model in MSQA Pipeline. 



\subsubsection{\textbf{Answer generation pipeline}}
The performance of our approach compared with the baselines are reported in Table \ref{tab:d5} which shows that our variants result in the highest performance in content preservation metrics such as ROGUE and BLEU. It is easier to generate answers for dichotomous questions compared to WH questions. Due to the subjective nature of WH questions, the reference answer given by a single user may not match the information passed as context to the model and subsequently will not match the generated response. Hence, performance metrics are higher for Dichotomous questions. 
In the absence of irrelevant information and ambiguous sentiment, the baseline models generate answers similar to our variants as evident from Question 1 in Table \ref{tab:d7}. When there are two or more questions, our variants tend to address the entire query while the baselines answer a single part of it. Questions 2 and 3 show MSQAP's capacity to handle conflicting sentiment compared to other variants and baselines due to sentiment filtering. 

\subsubsection{\textbf{Human Evaluation}}

\begin{table}
  \caption{Human evaluation of our methods}
  \label{tab:d6}
  \begin{tabular}{|l|ll|ll|}
    \cline{2-5}
    \multicolumn{1}{c|}{} & \multicolumn{2}{c|}{Dichot. questions} & \multicolumn{2}{c|}{WH questions} \\
    \cline{2-5}
    \multicolumn{1}{c|}{}& CC& CL& CC& CL\\
    \hline
    T5-QA & 0.919& 0.755& 0.833& 0.539\\
    MSQAP & \textbf{0.943} & \textbf{0.845}& \textbf{0.869}& \textbf{0.651}\\

    \hline
  \end{tabular}
\end{table}
\begin{table}
  \caption{User experience analysis}
  \label{tab:d10}
  \begin{tabular}{|l|l|}
    \cline{2-2}
    \multicolumn{1}{c|}{}& \% of questions\\
    \hline
    EP - Slightly better & 10.1\\
    EP - Much better  & 10.2\\
    MSQAP - Slightly better & \textbf{36.2}\\
    MSQAP - Much better  & \textbf{43.5}\\

    \hline
  \end{tabular}
\end{table}

The different variants of our pipeline outperform the baselines by a significant margin. But, minor difference in evaluation metrics between our variants is not enough to identify the best method. Hence, we performed human evaluation of generated answers and the results are tabulated in Table \ref{tab:d6}. Human annotators were asked to label a generated answer Correct w.r.t candidates (CC) if the information present in the context is accurately reflected in the generated response. The aim of this task is to evaluate if the model pays attention to information present in the candidates while generating the answers instead of memorizing the responses from training data. Correct w.r.t label (CL) is measured by checking if the generated response matches with the ground truth answer. Our pipeline generates answers with an overall accuracy of \textbf{77.2\%}. As discussed above, the drop in CL for WH questions is due to the subjective nature of reference answers even though the model is capturing the correct information from the context (CC). There is a slight improvement in CC in MSQAP when compared to the generation model (T5-QA) which we attribute to relevancy prediction. However, the huge improvement in CL can be attributed to both components. We also evaluated the user experience of our generation pipeline against an Extraction Pipeline (EP) currently in production which returns a single answer. We chose 842 questions where both the pipelines return the correct answer. We anonymized the pipeline names and shuffled the answers to avoid systemic bias. We asked the labelers to identify which pipeline answered their questions in a better way. Table \ref{tab:d10} summarizes the obtained labels and Table \ref{tab:d11} lists the answers obtained from the two pipelines. Labelers preferred our method for the first three questions and preferred extraction pipeline for the last two. In general, our approach answers the questions without unnecessary information which provides a better experience for annotators and lesser cognitive overload. When the extra information obtained from retrieval is useful for the question, annotators pick the extraction pipeline.
\begin{table}[h]
    \caption{Comparison of generated and retrieved answers}
      \label{tab:d11}
    \begin{tabular}{|p{8 cm} |}
        \hline

\textbf{Question 1: } Did I get a back cover of this mobile with the box\\
\textbf{EP: }Here's what you'll get in the box: Handset, Adapter (5V/2A), USB Cable, SIM Card Tool, Screen Protect Film, Important Info Booklet with Warranty Card, Quick Guide. 1 Year Warranty for Mobile and 6 Months for Accessories.  \\
\textbf{MSQAP: }No, you have to purchase it separately.  \\
\hline
\textbf{Question 2: } Is it best to play games?\\
\textbf{EP: } Realme 8 powerfull Mobile game and displays.\\
\textbf{MSQAP: } Yes, you can play games like pubg, asphalt 9 etc. \\
\hline
\textbf{Question 3: } how's camera quality good or bad\\
\textbf{EP: } The Primary Camera is 13MP + 8MP + 5MP + 2MP.\\
\textbf{MSQAP: } camera quality is good\\
\hline
\textbf{Question 4: } Is this phone has water resistance \\
\textbf{EP: } It Features IP53 Splash-proof Protection. \\
\textbf{MSQAP: } No, but it can survive a splash of water.\\
\hline
\textbf{Question 5: } Which displays is used led or amoled \\
\textbf{EP: } The display is a Full HD+ IPS Display. \\
\textbf{MSQAP: } It has a IPS LCD display\\
\hline
    \end{tabular}
\end{table}
\vspace{-1em}
\section{Conclusion}\label{conc}
Automatically answering questions is an important area of interest in E-commerce because it helps users to make better purchase decisions. We describe the challenges of using noisy and inconsistent user data. We propose a question answering pipeline (MSQAP) that utilizes information from different sources to generate coherent and accurate answers. Our relevancy prediction model (BERT-QA) outperforms all other variants and has an improvement of \textbf{12.36\%} in F1 score compared to the baseline. We also show how to solve for the ambiguity of sentiment in user data by using a pre-trained sentiment prediction model. We forgo the need for supervised annotations of text spans required in training Question Answering models and instead use answers submitted by users. Our generation pipeline has an average improvement of \textbf{37.43\%} in ROUGE and \textbf{198.31\%} in BLEU compared to the highest performing baseline (HSSC-q). Human evaluation of our pipeline shows an overall improvement in accuracy of \textbf{14.55\%} over the generation model (T5-QA) indicating accurate answers. We also show that our approach provides a better user experience for \textbf{79.7\%} of the questions. Our future work is to train an end-to-end model to incorporate both answer generation and sentiment prediction during training and also to extend this natural language generation approach to answering questions about offers, delivery, etc. in addition to product information.

\bibliographystyle{ACM-Reference-Format}
\bibliography{sample-base}


\end{document}